\documentclass[10pt,twocolumn,letterpaper]{article}

\usepackage[pagenumbers]{cvpr} 

\definecolor{darkred}{RGB}{192, 0, 0}
\definecolor{darkgreen}{RGB}{0, 100, 0}
\usepackage{multirow}
\usepackage[numbers,sort&compress]{natbib}

\usepackage{graphicx}
\usepackage{makecell}
\definecolor{cvprblue}{rgb}{0.21,0.49,0.74}
\usepackage[pagebackref,breaklinks,colorlinks,allcolors=cvprblue]{hyperref}

\def\paperID{***} 
\def\confName{CVPR}
\def\confYear{2026}

\title{UniRSCD: A Unified Novel Architectural Paradigm for Remote Sensing Change Detection}

\author{Yuan Qu$^1$, Zhipeng Zhang$^2$\thanks{Yuan Qu and Zhipeng Zhang contributed equally to this work.}, Chaojun Xu$^1$, Qiao Wan$^3$, Mengying Xie$^4$, Yuzeng Chen$^5$,\\
	Zhenqi Liu$^1$\thanks{Corresponding author}, Yanfei Zhong$^3$\\
	$^1$College of Artificial Intelligence, Southwest University\\ 
	$^2$School of Artificial Intelligence, Shanghai Jiaotong University\hspace{4pt}$^3$LIESMARS, Wuhan University\\
	 $^4$CS, Chongqing University\hspace{4pt}$^5$School of Geodesy and Geomatics, Wuhan University\\
	{\tt\small zhipeng.zhang.cv@outlook.com, liuzhenqi@swu.edu.cn, zhongyanfei@whu.edu.cn}
}

\begin{document}
\maketitle
\begin{abstract}
In recent years, remote sensing change detection has garnered significant attention due to its critical role in resource monitoring and disaster assessment. Change detection tasks exist with different output granularities such as BCD, SCD, and BDA. However, existing methods require substantial expert knowledge to design specialized decoders that compensate for information loss during encoding across different tasks. This not only introduces uncertainty into the process of selecting optimal models for abrupt change scenarios (such as disaster outbreaks) but also limits the universality of these architectures. To address these challenges, this paper proposes a unified, general change detection framework named UniRSCD. Building upon a state space model backbone, we introduce a frequency change prompt generator as a unified encoder. The encoder dynamically scans bitemporal global context information while integrating high-frequency details with low-frequency holistic information, thereby eliminating the need for specialized decoders for feature compensation. Subsequently, the unified decoder and prediction head establish a shared representation space through hierarchical feature interaction and task-adaptive output mapping. This integrating various tasks such as binary change detection and semantic change detection into a unified architecture, thereby accommodating the differing output granularity requirements of distinct change detection tasks. Experimental results demonstrate that the proposed architecture can adapt to multiple change detection tasks and achieves leading performance on five datasets, including the binary change dataset LEVIR-CD, the semantic change dataset SECOND, and the building damage assessment dataset xBD.
\end{abstract}
    
\section{Introduction}
\label{sec:intro}

In recent years, remote sensing change detection has garnered widespread attention due to its crucial role in resource monitoring, environmental assessment, and disaster response. With the advancement of deep learning, various methods have been proposed to address diverse application needs, such as binary change detection \cite{XU2026112878,YANG2026112169,9883686}, semantic change detection \cite{10419228,10543161,10574229}, and time-series analysis \cite{10384241,zhu2025change3drevisitingchangedetection}.\par
\begin{figure*}[!t]
	\centering
	\includegraphics[scale=0.40,trim=20pt 30pt 30pt 30pt,clip]{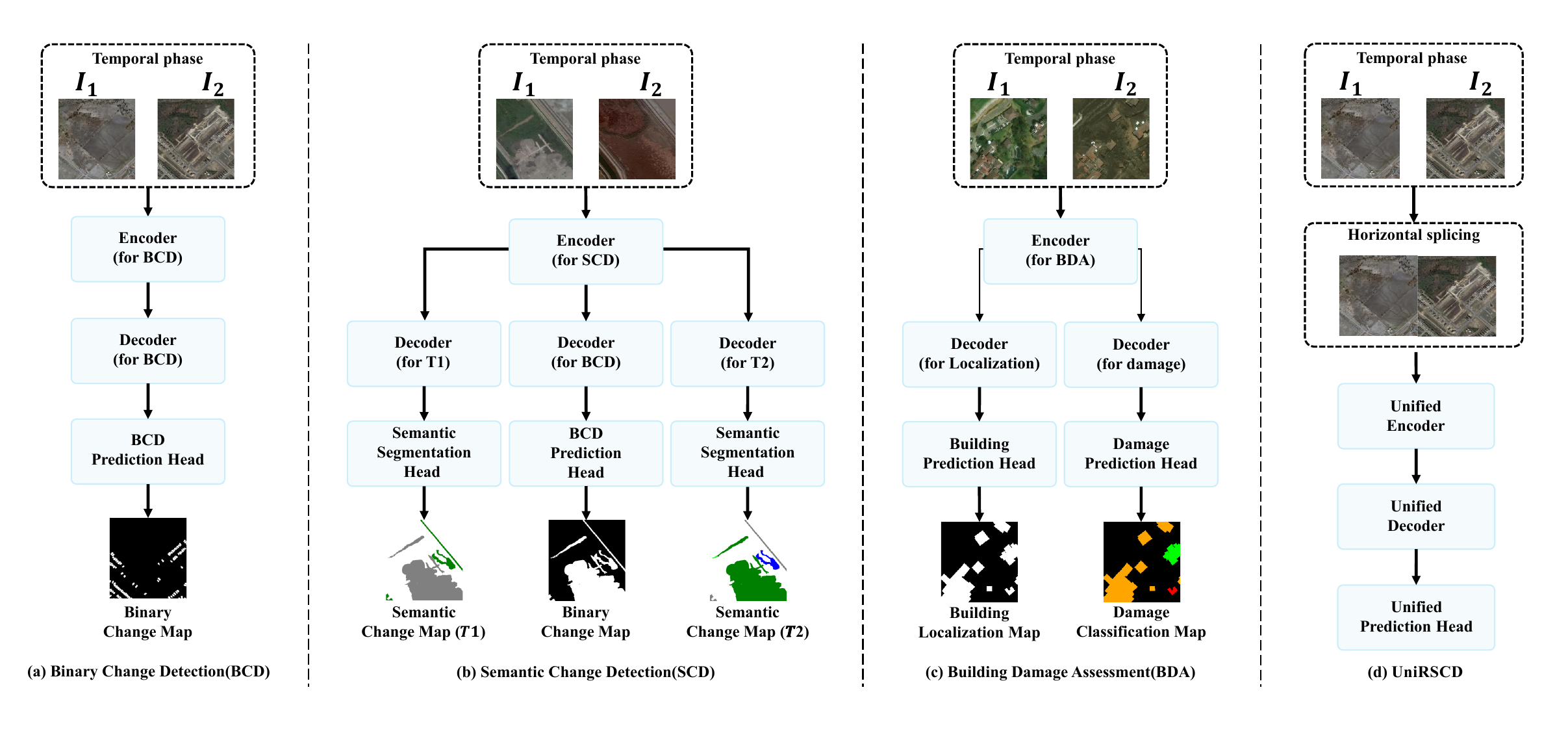}
	\caption{Overall architectures for BCD task, SCD task, BDA task and UniRSCD. (a) BCD task obtains features representing changed objects. (b) SCD task obtains features representing semantic changes and binary changes, where the T1 and T2 decoders share the same prediction heads. (c) BDA task: Two features used for building localization and damage classification. (d) Our UniRSCD uses horizontal concatenation to process bitemporal images, then handles multiple tasks through a unified architecture.}
	\label{fig1}
\end{figure*}
In the field of remote sensing change detection, inherent differences exist in feature interaction patterns and output granularity across various tasks, such as binary change detection \cite{10419228,9883686,10221754}, semantic change detection \cite{10443352,NEURIPS2023_788e086c}, and building damage assessment \cite{ZHENG2021112636,9883686,zhu2025change3drevisitingchangedetection}. As shown in Figure \ref{fig1}, in order to follow the task-driven design paradigm, existing methods require custom decoders \cite{10565926} to compensate for information lost during encoding. These methods are built upon two primary neural network architectures: convolutional neural networks (CNNs) \cite{10005110,10368086,10540174} and visual transformers \cite{10623172}. CNN-based approaches effectively capture local spatial features through convolutional inductive bias \cite{WU2025111010,8451652}, yet their limited receptive fields restrict modeling of long-range dependencies. In contrast, transformers utilize self-attention to capture global information \cite{10493070,9491802,9883686}, but their uniform weighting often smooths out fine-grained details, reducing the model's focus on granular information. Therefore, existing methods employ customized decoders to compensate for information loss during the encoding phase, thereby addressing the aforementioned issues. However, this custom design hinders the development of a unified framework, thereby preventing the consolidation of diverse change detection tasks with varying output granularities under a unified architecture.
\par To address the above problems, this paper proposes a generalized change detection framework with a single-stream architecture. This framework constructs its input by horizontally concatenating bitemporal images and utilizes state space models to capture long-range spatiotemporal dependencies \cite{gu2022efficientlymodelinglongsequences} from shallow to deep layers. Meanwhile, the framework introduces a frequency change prompt generator module that implicitly formulates change detection as a two-stage optimization process: first extracting low-frequency structures to locate candidate change regions, then integrating high-frequency details to refine boundaries. The design combining state space models with frequency prompt compensates for information loss during encoding in existing methods, eliminating the need to build task-specific decoders for different change detection tasks. Additionally, the framework also designs a unified decoder and prediction head, which unifies multiple tasks into a shared representation space through hierarchical feature interaction and task-adaptive output mapping, thereby constructing a unified change detection framework. \par
Experiments demonstrate that UniRSCD achieves an $IoU$ score of 72.03 on the SYSU binary change detection dataset, an $mIoU$ score of 72.7 on the SECOND semantic change detection dataset, and an $F_1^{overall}$ score of 78.02 on building damage assessment dataset. Benefiting from its single-stream architecture and simplified decoder design, UniRSCD significantly reduces training time compared to existing methods, enabling rapid and precise detection of sudden change scenarios. In summary, the main innovations of this paper are reflected in the following three aspects:
\begin{itemize}
	\item We integrate a state space model with a frequency change prompt generator module to effectively capture both global contextual information and local detailed features, thereby eliminating the need for specialized decoders to compensate for information loss during encoding.
	\item We designed a simple decoder and a multi-task adaptive prediction head to unify various change detection tasks.
	\item We conduct experiments on three types of tasks (Binary Change Detection, Semantic Change Detection, and Building Damage Assessment). The framework can quickly adapt to different tasks and achieves state-of-the-art (SOTA) results on the datasets.
\end{itemize}

\section{Related Work}
\label{sec:Related Work}

\subsection{Deep Learning Change Detection}

Current change detection communities have developed various architectures for different change detection tasks. For example, it has branched into Binary Change Detection for identifying surface changes \cite{10384241,XU2026112878,YANG2026112169,10755134}, Semantic Change Detection for localizing and categorizing changes \cite{10422819,10543161,10574229,10493070}, and Building Damage Assessment for classifying building damage levels post-disaster \cite{10565926,zhu2025change3drevisitingchangedetection}. Current mainstream change detection approaches primarily rely on convolutional neural networks (CNNs) \cite{10005110,10368086,10540174,WU2025111010} and transformer architectures \cite{LUO2026131874,LIU2026137,LI2024103663,10623172,10493070}. CNNs excel at extracting local spatial features through their local connections and inductive biases \cite{8451652,NEURIPS2023_788e086c,10755134,10011164}. Conversely, visual transformers effectively capture global contextual information via self-attention mechanisms \cite{10493070,xu2024hybrid}. Although the above methods have developed different decoders and prediction heads for various tasks to compensate for information loss during the encoding process, achieving high accuracy, this approach hinders the development of a unified architecture and also limits the adaptability and deployment efficiency of the model. We propose a unified encoding architecture that integrates the state space model (SSM) with a frequency change prompt generator module. This approach ensures both global consistency and local precision of features during the encoding phase, thereby eliminating the need for complex task-specific decoders. It enables high-performance detection across BCD, SCD, and BDA under one architecture, enhancing integration and scalability.

\subsection{State Space Models}

In recent years, State Space Models (SSMs) have provided new perspectives for remote sensing change detection (CD) due to their remarkable efficiency in long-sequence modeling. Particularly, architectures represented by Mamba \cite{10565926}, leveraging their linear time complexity and selective scanning mechanism, can efficiently capture global spatiotemporal dependencies in multitemporal remote sensing images, significantly enhancing the efficiency of processing high-resolution imagery and the capability of feature representation \cite{2024arXiv240109417Z,10565926,LIU202573,smith2023convolutionalstatespacemodels,liu2024vmambavisualstatespace,11094549,shi2025super,10.1117/1.JRS.19.046507,XU2025104364}. However, despite the significant progress these methods have achieved in computational performance and feature modeling, they generally lack specialized suppression mechanisms designed for the inherent pseudo-changes in remote sensing CD, such as apparent land surface changes induced by seasonal variations, illumination differences, registration errors, or sensor noise \cite{9849109,LIU2026137,10445290,LI2024103663}. Consequently, although existing SSM-based methods possess advantages in efficiency and feature extraction, their robustness against pseudo-changes remains a key limitation and a core challenge to be addressed when dealing with complex real-world scenarios.

\section{Method}

\subsection{Overview}

\begin{figure*}[t]
	\centering
	\includegraphics[scale=0.58,trim=79pt 0pt 60pt 0pt,clip]{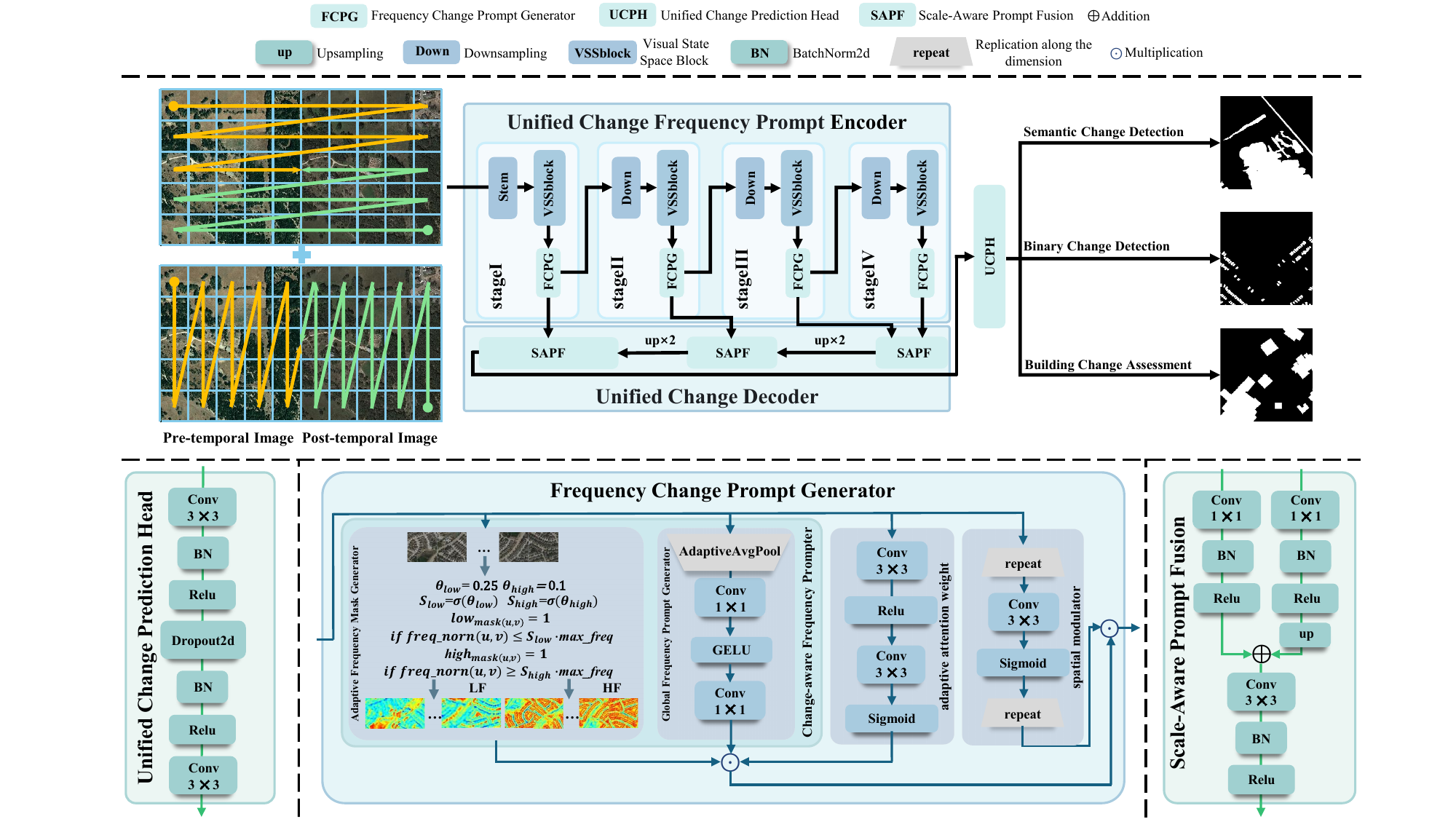}
	\caption{The overall architecture of UniRSCD. The upper half of the figure illustrates the main workflow of UniRSCD. Horizontally concatenated bitemporal images are scanned along row and column directions and their inverses for input. Data flows sequentially to the right through processing units from Stage I to Stage IV, which include frequency change prompt generator (FCPG) and VSSBlock modules. After passing through the SAPF module and undergoing upsampling, the data is directed via UCPH to three parallel task branches: semantic change detection, binary change detection, and building change assessment. The lower half of the figure details the internal structure of the Frequency Change Prompt Generator (FCPG), Unified Change Prediction Head (UCPH), and Scale-Aware Prompt Fusion (SAPF).}
	\label{fig2}
\end{figure*}

Based on state space models \cite{10565926,LIU202573,2024arXiv240109417Z}, we propose UniRSCD, a unified change detection framework with a single-stream architecture \cite{10.1007/978-3-031-20047-2_20} that enables multi-task processing through a single decoder and prediction head. The core design of this architecture integrates a frequency prompt mechanism into the state space mode. This allows a single framework to handle multiple change detection tasks without requiring dedicated decoders for each task. As illustrated in Figure \ref{fig2}, UniRSCD comprises three modules: a unified change frequency prompt encoder (built upon a state space model backbone and introducing a frequency change prompt generator), a unified change decoder, and a unified prediction head.

\subsection{Unified Change Frequency Prompt Encoder}
The unified change frequency prompt encoder consists of a horizontally concatenated bitemporal images input, a 4-stage backbone network based on state space model, frequency change prompt generator integrated at each stage, and multi-scale feature outputs, forming a unified encoding architecture specifically designed for remote sensing change detection.\par
\textbf{Horizontal Concatenation.} Given the registered bitemporal images $X_{pre},X_{post}\in R^{C \times H \times W}$, where $C$ is the number of input channels, and \textit{H} and \textit{W} are the spatial dimensions, we perform a horizontal concatenation along the width dimension, as:
\begin{equation}\begin{aligned}
		X_{concat}&=Concat_{width}\left(X_{pre},X_{post}\right)
		\label{eq1}
\end{aligned}\end{equation}
This concatenation preserves spatiotemporal layout and correspondence. We then employ a four-directional cross-scanning strategy to convert the 2D feature map $X_{concat}$ into sequential representations. Specifically, we scan along the row, column, and their reverse directions, flattening the 2D features into a 1D sequence $S_{d}\in R^{L\times C}$, where $L=H\times2W$ denotes the sequence length. This multi-directional scanning captures cross-temporal features from diverse perspectives, offering comprehensive spatial context for change detection. The generated sequences are processed and then restored to spatial features via reverse scanning as $F_d=Reshape_d(S_d^{\prime})$, where $S_{d}^{\prime}$ is the processed sequence. Finally, features from all four directions are aggregated to form a unified multi-scale representation. \par
\textbf{Unified Change Frequency Prompt Encoder.} Our change frequency prompt backbone network consists of a state space model backbone network and a frequency change prompt generator.\par \textit{1) State Space Model Backbone.} We adopt the Selective State Space Model \cite{gu2022efficientlymodelinglongsequences} as the core building block to model complex spatiotemporal dependencies in bitemporal features. Given an input sequence representation $x\in R^{L \times D}$, where $L$ is the sequence length and $D$ is the feature dimension, the state space model achieves information propagation and integration through the hidden state $h_{t}\in R^{N}$. Its continuous-time form is expressed as:
\begin{equation}
	\begin{aligned}
		y \left( t \right)&=Ch \left( t \right)+Dx \left( t \right),\\
		h^{\prime}(t)&=Ah(t)+Bx(t),
		\label{eq2}
\end{aligned}\end{equation}
the selective state space model employs input-dependent parameters for B, C, and $\Delta$, which are generated from the input $X_{t}$ via linear projections as $\Delta{=softplus(w_{\Delta x_{t}}),B =W_{B}x_{t},C =W_{C}x_{t}}$, where $w_{ \Delta},w_{B},w_{C}$ are learnable weights. This adaptive parameterization enhances sensitivity to subtle changes by enabling the model to focus on relevant information while ignoring redundancies. The visual state space block integrates the state space branch and the MLP branch into a unified architecture. The state space branch uses layer normalization, a selective state space model, and deep convolutions to model global dependencies and local features. The \textit{MLP} branch employs fully connected layers and \textit{GELU} for nonlinear transformation. Outputs from both branches are fused with the inputs via stochastic depth, forming a residual learning framework that captures both global patterns and local interactions, thereby providing multi-scale hierarchical features for change detection. \par
\textit{2) Frequency Change Prompt Generator.} Inspired by frequency-selective processing in human vision \cite{9849109,LIU2026137}, we design a frequency change prompt generator (FCPG) to enhance change-related features in the frequency domain. For an input feature $X\in R^{C\times H\times W}$, the FCPG computes the radial frequency distribution $F_{radial}=\sqrt{F_{h}^{2}+F_{w}^{2}}$, where $F_{h}$ and $F_{w}$ are the frequency components along the height and width. Learnable thresholds are then used to construct low-and high-frequency masks. As shown in Figure \ref{fig2}, each frequency band first passes through a global frequency prompt generator for base weights, followed by an adaptive attention mechanism for spatial refinement. The final prompt map for each band is obtained via a triple-product operation: 
\begin{equation}
	P_b=W_b^{global}\odot M_b\odot W_b^{spatial}.
	\label{eq3}
\end{equation}
The separated frequency band features are then reintegrated into the original feature space using a feature fusion convolution. To preserve spatial continuity, a Spatial Modulator (\textit{SPM}) is applied to enhance local consistency, producing the modulated output as $P_{modulated}=P_{fused}\odot SPM(P_{grouped})$, where $X_{grouped}$ represents the mean features after channel grouping. The modulated frequency prompts are fused with the original features via a residual connection, as $X_{output}=X+\alpha\cdot P_{modulated}$, where $\alpha$ is an adjustable fusion coefficient. This design integrates global statistics, local frequency bands, and spatial weights, effectively fusing multi-dimensional information. By explicitly modeling in the frequency domain, the FCPG module robustly distinguishes real object changes from pseudo-changes, enhancing detection performance in complex scenarios. 

\subsection{Unified Change Decoder and Prediction Head}

As shown in Figure \ref{fig2}, UniRSCD employs a simple feature pyramid network as its decoder and a prediction head for multiple change detection tasks.\par
\textbf{Unified Change Decoder.} We design a unified change decoder based on a feature pyramid network to integrate multi-scale encoder features $F^{1},F^{2},F^{3},F^{4}$ via top-down propagation. First, features from each level are projected into a unified space using 1×1 convolutions. Then, a top-down fusion process starts from the highest level, progressively upsampling and merging features with lower levels as formulated by:
\begin{equation}
	N_{\mathrm{i}}=Conv_{3\times3}(Up(N_{\mathrm{i+1}})\oplus P_{\mathrm{i}}),
	\label{eq4}
\end{equation}
where \textit{Up} denotes the bilinear upsampling operation and $\oplus$ represents element-wise addition. To enhance feature discriminability, we introduce skip connections with 3×3 convolutions at each fusion stage. The final change representation is then constructed by concatenating and refining the upsampled multi-scale features, as $C=Conv_{3\times3}(Concat(Up(N_4),Up(N_3),Up(N_2),N_1))$, where \textit{Up} denotes upsampling. This multi-scale fusion strategy leverages both high-level semantics and low-level details, thereby enriching context for change prediction while improving boundary localization and semantic consistency.\par
\textbf{Unified Prediction Head.} We design a unified prediction head that achieves unified support for various tasks, including Binary Change Detection (BCD), Semantic Change Detection (SCD), and Building Damage Assessment (BDA), by task-specific adaptation. Given an input feature map $X\in R^{C_{in}\times H\times W}$, the unified feature extraction process can be formulated as: 
\begin{equation}\begin{aligned}
		Z_{1} & =Dropout(\sigma(BN(Conv_{3\times3}(X,W_{1})))),\\
		Z_{2} & =\sigma(BN(Conv_{3\times3}(Z_{1},W_{2}))),
		\label{eq5}
\end{aligned}\end{equation}
where $\mathrm{Conv}_{3\times3}$ is a 3×3 convolution, BN is Batch Normalization, $\sigma$ is a configurable activation function (\textit{ReLU}), and random feature dropping is applied with a probability $p=0.2$. This design progressively reduces feature dimensions to $C_{hid}/2$, balancing capacity and complexity. For multi-task output: the BCD task uses a single convolutional layer; the SCD task employs three parallel output heads for change detection and bitemporal semantic segmentation, enhanced through background suppression mechanisms, thereby improving the ability to distinguish foreground from background. The BDA task utilizes a dual-branch design for simultaneous building segmentation and damage assessment. This unified framework supports multiple change detection tasks, ensuring practicality and extensibility.

\subsection{Multi-Task Adaptive Loss Function}
During the training of change detection models, the appropriate loss function is automatically selected based on different task types. These loss functions are designed as a combination of cross-entropy loss and Lovasz loss, ensuring both category balance optimization and optimization for the IoU metric. \par
For Binary Change Detection, we design a loss function formulation to supervise the change detection task. We adopt a combination of cross-entropy loss and Lovasz loss to handle pixel-level accuracy and segmentation quality. The final loss is defined as:
\begin{equation}
	L_{bcd}=L_{ce}+0.75\times L_{lovasz},
	\label{eq6}
\end{equation}
where $L_{ce}$ provides category-balanced optimization, while $L_{lovasz}$ directly optimizes the IoU metric, effectively addressing the imbalanced characteristics of change detection datasets.\par
For Building Damage Assessment, we extend the formulation (\ref{eq6}) to simultaneously optimize building localization and damage classification. The multi-task loss integrates:
\begin{equation}
	L_{bda}=L_{cc}^{loc}+L_{cc}^{clf}+0.5\times L_{lovasz}^{loc}+L_{lovasz}^{clf}.
	\label{eq7}
\end{equation}
$L_{cc}^{loc}$ and $L_{cc}^{clf}$ provide category-balanced supervision for building localization and damage classification, respectively, while $L_{lovasz}^{loc}$ and $L_{lovasz}^{clf}$ directly optimize the IoU metrics for localization and classification. \par
For Semantic Change Detection, we introduce a complex triple-branch supervision scheme, which includes change detection, bitemporal semantic segmentation, and temporal consistency. The loss function comprises:
\begin{equation}\begin{aligned}
		L_{scd} & =L_{cc}^{cd}+0.5\times(L_{cc}^{t1}+L_{cc}^{t2}+0.5\times L_{sim})+\\
		&0.75\times(L_{lovasz}^{cd}+0.5\times(L_{lovasz}^{t1}+L_{lovasz}^{t2})).
		\label{eq8}
\end{aligned}\end{equation}
in this formulation, $L_{cc}^{cd}$, $L_{cc}^{t1}$, and $L_{cc}^{t2}$ provide category-balanced supervision for change detection, semantic segmentation at time one, and semantic segmentation at time two, respectively, while $L_{lovasz}^{cd}$, $L_{lovasz}^{t1}$, and $L_{lovasz}^{t2}$ optimize the corresponding IoU metrics for each task. 

\section{Experiments and Results}
\subsection{Experimental Setup}

\textbf{Datasets and Evaluation.} Metrics: We conduct a comprehensive evaluation on five public change detection benchmark datasets: LEVIR-CD \cite{chen2020spatial}, WHU-CD \cite{ji2018fully}, SYSU \cite{shi2021deeply}, SECOND \cite{yang2021asymmetric}, and xBD \cite{gupta2019creating}. LEVIR-CD, WHU-CD, and SYSU are three binary change detection (BCD) datasets, containing changes in buildings, roads, and lakes. SECOND is a semantic change detection (SCD) dataset with six land cover categories for semantic learning. xBD is a building damage assessment (BDA) dataset providing information on four damage levels. For the change detection task, we adopt Precision, Recall, F1-score, and Intersection over Union (IoU); for semantic change detection, we use Overall Accuracy (OA), mean Intersection over Union (mIoU), Semantic Kappa coefficient (SeK), and the F1 score; for disaster assessment, we employ the localization F1-score, damage classification F1-score, and overall F1-score.\par
\textbf{Implementation Details.} In this study, all experiments were conducted on a hardware platform equipped with an NVIDIA RTX 5880 Ada GPU, utilizing the PyTorch 2.0 framework and CUDA 11.8 for accelerated computing. The proposed unified architecture employs the vmamba's small version \cite{liu2024vmamba}. For the BCD task, the training parameters were uniformly set as follows: batch size of 22, input image crop size of 256×256, maximum iterations of 320,000, initial learning rate of 1e-4, and weight decay of 5e-4. For the BDA task, the training parameters were set to: batch size of 16, input image crop size of 256×256, maximum iterations of 400,000, initial learning rate of 1e-4, and weight decay of 5e-4. For the SCD task, the parameters were: batch size of 16, input image crop size of 256×256, maximum iterations of 400,000, initial learning rate of 1e-4, and weight decay of 5e-4. Random rotation, left-right flipping, and top-bottom flipping were used as training data augmentation methods.\par
This study employs a unified two-stage training strategy. In the first stage, all modules except the frequency change prompt generator are trained using full learning rates and weight decay to learn robust spatiotemporal features. In the second stage, the backbone network is frozen, and only the frequency change prompt generator and prediction head are activated and trained using a lower learning rate (1e-5) to refine details. The optimization process employs the AdamW optimizer \cite{loshchilov2017decoupled} with a StepLR learning rate scheduler, integrating multi-task losses through weighted summation. 

\subsection{Comparison with State-of-the-Art Methods}

\begin{table}[t]
	\centering
	\caption{The performance comparison of different binary change detection methods on the SYSU dataset. The best, second-best, and third-best results are highlighted in red, blue, and green, respectively. All results for the three evaluation metrics are expressed as percentages (\%).} 
	\label{tab1}
	\begin{tabular}{ccccc} 
		\toprule
		Method & Rec & Pre & F1 & IoU \\
		\midrule
		BIT \cite{9491802} & 75.56 & 83.59 & 79.44 & 65.94 \\
		SEIFNet \cite{10419228} & 78.35 & 83.59 & 80.88 & 67.9 \\
		ChangeFormer \cite{9883686} & 78.51 & 77.16	& 77.83	& 63.71	\\
		AFCF3D-Net \cite{10221754} & \textcolor{darkred}{83.88} & 82.3 & 83.11 &	71.09  \\
		AACRNet \cite{LI2026105594} & \textcolor{blue}{82.62} & 84.55	 & \textcolor{blue}{83.57}	& \textcolor{blue}{71.78} \\
		ChangeMamba \cite{10565926} & 77.82 & \textcolor{darkred}{87.89} &	82.55 & 70.28  \\
		Change3D \cite{zhu2025change3drevisitingchangedetection} & \textcolor{darkgreen}{81.39} & \textcolor{darkgreen}{84.89} & 83.11 & \textcolor{darkgreen}{71.1}	\\
		UniRSCD & 80.62	& \textcolor{blue}{87.11}	& \textcolor{darkred}{83.74}	& \textcolor{darkred}{72.03} \\ 
		\bottomrule
	\end{tabular}
\end{table}

To validate the effectiveness of the proposed method, we conduct comprehensive comparisons with various representative state-of-the-art (SOTA) methods in the field of remote sensing change detection in recent years, including BIT \cite{9491802}, SEIFNet \cite{10419228}, ChangeFormer \cite{9883686}, AFCF3D-Net \cite{10221754}, AACRNet \cite{LI2026105594}, ChangeMamba \cite{10565926}, Change3D \cite{zhu2025change3drevisitingchangedetection}, HRSCD-S4 \cite{daudt2019multitask}, ChangeMask \cite{ZHENG2022228}, SSCD \cite{NEURIPS2023_788e086c}, BiSRNet \cite{NEURIPS2023_788e086c}, TED \cite{10443352}, xBD baseline \cite{Gupta_2019_CVPR_Workshops}, MTF \cite{weber2020building}, ChangeOS-R50 \cite{zheng2021building}, ChangeOS-R101 \cite{zheng2021building}, DamFormer \cite{chen2022dual}. All comparative methods are reproduced based on their official codebases or directly cite the best reported results.\par

\begin{table}[!t]
	\centering
	\caption{The performance comparison of different binary change detection methods on the LEVIR-CD dataset.}
	\label{tab2}
	\begin{tabular}{ccccc} 
		\toprule
		Method & Rec & Pre & F1 & IoU \\
		\midrule
		BIT \cite{9491802} & 89.37 & 89.24 & 89.31 & 80.68 \\
		SEIFNet \cite{10419228} & 88.28 & 91.74 & 89.98 & 81.78 \\
		ChangeFormer \cite{9883686} & 88.8 & 92.05	& 90.4 & 82.48	\\
		AFCF3D-Net \cite{10221754} & 90.17 & 91.35 & 90.76 &	83.08 \\
		AACRNet \cite{LI2026105594} & \textcolor{blue}{91.68} & \textcolor{blue}{92.69}	 & \textcolor{darkgreen}{92.18}	& \textcolor{darkgreen}{85.5} \\
		ChangeMamba \cite{10565926} & \textcolor{darkgreen}{91.67} & \textcolor{darkred}{92.72} &	\textcolor{blue}{92.19} & \textcolor{blue}{85.52}  \\
		Change3D \cite{zhu2025change3drevisitingchangedetection} & 91.64 & 91.58 & 91.82 & 84.87	\\
		UniRSCD & \textcolor{darkred}{92.87}	& \textcolor{darkgreen}{91.78}	& \textcolor{darkred}{92.28}	& \textcolor{darkred}{85.67} \\ 	
		\bottomrule
	\end{tabular}
\end{table}
\begin{table}[!t]
	\centering
	\caption{The performance comparison of different binary change detection methods on the WHU-CD dataset.
	} 
	\label{tab3}
	\begin{tabular}{ccccc} 
		\toprule
		Method & Rec & Pre & F1 & IoU \\
		\midrule
		BIT \cite{9491802} & 81.48 & 86.64 & 83.98 & 72.39 \\
		SEIFNet \cite{10419228} & 72.55 & 87.73 & 79.42 & 65.87 \\
		ChangeFormer \cite{9883686} & 90.28 & 89.36 & 89.82 &	81.6 \\
		AFCF3D-Net \cite{10221754} & \textcolor{blue}{93.69} & 93.47	& 93.58	 & 87.93  \\
		AACRNet \cite{LI2026105594} & 93.31 & 95.23 &	\textcolor{darkgreen}{94.26} & \textcolor{darkgreen}{89.14} \\
		ChangeMamba \cite{10565926} & 92.29 &  \textcolor{blue}{95.9} & 94.06 &  88.79 \\
		Change3D \cite{zhu2025change3drevisitingchangedetection} &  \textcolor{darkgreen}{93.5} & \textcolor{darkred}{96.33} & \textcolor{blue}{94.57} & \textcolor{blue}{89.71} \\
		UniRSCD & \textcolor{darkred}{94.08}	& \textcolor{darkgreen}{95.55}	& \textcolor{darkred}{94.81}	& \textcolor{darkred}{90.14} \\ 
		\bottomrule
	\end{tabular}
\end{table}
\textbf{Binary Change Detection.} UniRSCD achieves an F1 score 0.63\% higher than AFCF3D-Net on the SYSU dataset as shown in Table \ref{tab1}. On the LEVIR-CD dataset as shown in Table \ref{tab2}, its F1 and IoU surpass ChangeMamba-Small by 0.09\% and 0.15\%, respectively. On the WHU-CD dataset as shown in Table \ref{tab3}, it achieved comparable performance to state-of-the-art methods, demonstrating excellent general detection capabilities. \par
\begin{table}[!t]
	\centering
	\caption{The performance comparison of different Semantic Change Detection methods on the SECOND dataset.} 
	\label{tab4}
	\begin{tabular}{ccccc} 
		\toprule
		Method & OA & F1 & mloU & SeK37 \\
		\midrule
		HRSCD-S4 \cite{daudt2019multitask} & 83.88 & 47.69 & 70.58 & 16.81 \\
		ChangeMask \cite{ZHENG2022228} & 85.53 & 50.54 & 71.39 & 18.36 \\
		SSCD \cite{NEURIPS2023_788e086c} & 85.62 & 52.8 & 72.14 & 20.15 \\
		BiSRNet \cite{NEURIPS2023_788e086c} & \textcolor{darkgreen}{85.8} & \textcolor{darkgreen}{53.59} & \textcolor{blue}{72.69} &	\textcolor{blue}{21.18} \\
		TED \cite{10443352} & 85.5 & 52.61	 & \textcolor{darkgreen}{72.19}	&  20.29 \\
		Change3D \cite{zhu2025change3drevisitingchangedetection} & \textcolor{darkred}{86.78} & \textcolor{darkred}{61.25}	 & 71.78	& \textcolor{blue}{20.89} \\
		UniRSCD & \textcolor{blue}{86.44}	& \textcolor{blue}{56.35}	& \textcolor{darkred}{72.7}	& \textcolor{darkred}{22.56} \\ 	
		\bottomrule
	\end{tabular}
\end{table}

\begin{table*}[t]
	\centering
	\caption{The performance comparison of different Building Damage Assessment methods on the xBD dataset.}
	\begin{tabular}{lccccccc}
		\toprule
		\multirow{2}{*}{Method} & \multirow{2}{*}{$F_1^{loc}$} & \multirow{2}{*}{$F_1^{clf}$} & \multirow{2}{*}{$F_1^{overall}$}  & \multicolumn{4}{c}{Damage $F_1$ per-class}\\
		\cmidrule(r){5-8}
		&  &  &  & No damage & Minor damage & Major damage & Destroyed\\
		\midrule
		xBD baseline \cite{Gupta_2019_CVPR_Workshops}	 & 80.47	& 3.42	 & 26.54 & 66.31 & 14.35	& 0.94	 & 46.57 \\
		MTF \cite{weber2020building} & 83.6	& 70.02	& 74.1	& \textcolor{darkgreen}{90.6} & 49.3	& 72.2 & 	83.7 \\
		ChangeOS-R50 \cite{zheng2021building} & \textcolor{darkgreen}{85.41} &	70.88	& 75.24		& 88.98	& 53.33 & 71.24 & 80.6 \\
		ChangeOS-R101 \cite{zheng2021building} & \textcolor{blue}{85.69}	& 71.14 & \textcolor{darkgreen}{75.5}	& 89.11	& 53.11	& 72.44	& \textcolor{darkgreen}{80.79} \\
		DamFormer \cite{chen2022dual} & \textcolor{darkred}{86.86}	& \textcolor{darkgreen}{72.81}	& \textcolor{blue}{77.02}	& 89.86	& \textcolor{darkred}{56.78}	& \textcolor{darkgreen}{72.56} & 80.51\\
		Change3D \cite{zhu2025change3drevisitingchangedetection} & 79.86	& \textcolor{blue}{73.25}	& 74.99	& \textcolor{darkred}{98.75}	& \textcolor{darkgreen}{53.7}	& \textcolor{blue}{73.72}	& \textcolor{blue}{84.7} \\
		UniRSCD & 84.47	& \textcolor{darkred}{75.26}	& \textcolor{darkred}{78.02}		& \textcolor{blue}{94.76}	& \textcolor{blue}{56.64}	& \textcolor{darkred}{75.71}	& \textcolor{darkred}{85.22} \\
		\bottomrule
	\end{tabular}
	\label{tab5}
\end{table*}

\textbf{Semantic Change Detection.} UniRSCD demonstrated exceptional performance on the SECOND dataset as shown in Table \ref{tab4}. UniRSCD surpassed ChangeMask by 5.81\% in F1 score and 4.20\% in semantic Kappa, while outperforming Change3D by 1.67\% in SeK and 0.92\% in mIoU, achieving optimal results on most metrics and highlighting its superior capability in complex semantic understanding. \par

\textbf{Building Damage Assessment.} As shown in Table \ref{tab5}, UniRSCD achieved an overall F1 score exceeding DamFormer by 1.00\% on the xBD dataset, with a remarkable 85.22\% precision for detecting “destroyed” damage levels. UniRSCD outperformed Change3D in overall F1 scores by margins of 3.03\%, further validating its multi-task adaptability. \par
These improvements demonstrate the effectiveness of the UniRSCD unified single-stream Mamba architecture and frequency change prompt module. The former effectively captures long-range spatiotemporal dependencies through a selective state space model, fully interacting with bitemporal features. The latter suppresses pseudo-variations caused by seasonal changes and sensor differences by employing candidate region localization guided by low-frequency holistic information and boundary optimization driven by high-frequency detail information.
\subsection{Diagnostic Study}
To systematically verify the necessity and synergy of each core component in the UniRSCD framework, we conducted comprehensive ablation experiments on the SYSU dataset. All experiments employed identical training settings to ensure fair comparisons. \par
\begin{table}[!t]
	\centering
	\caption{Validation of the effectiveness of different modules of UniRSCD on the SYSU dataset. “PrM” is pretrained model, “FCPG” is frequency change prompt generator, “AAF” is adaptive adjustment of frequency settings, “SPM” is spatial modulator, and “HLF” is high and low frequencies.} 
	\label{tab6}
	\setlength{\tabcolsep}{2.9pt}
	\begin{tabular}{ccccccccc} 
		\toprule
		PrM&FCPG&SPM&HLF&AAF&Res&Pre&F1&IoU\\
		\midrule
		& \checkmark & \checkmark & \checkmark & \checkmark & 76.56 & 80.09 & 78.29 & 71.79 \\ 
		\checkmark &   &   &   &   & 79.11 & 86.78 & 83.42 & 71.56 \\
		\checkmark & \checkmark &   &   &   & 82.45 & 85.23 & 83.07 & 71.52 \\
		\checkmark & \checkmark & \checkmark &   &   & 82.31 & 85.74 & 83.14 & 71.67 \\
		\checkmark & \checkmark & \checkmark & \checkmark &   & 79.76 & 86.54 & 83.16 & 71.23 \\
		\checkmark & \checkmark & \checkmark & \checkmark & \checkmark & 80.62 & 87.11 & 83.74 & 72.03 \\
		\bottomrule
	\end{tabular}
\end{table}
\textbf{Effectiveness of Architecture Design.} We first evaluated the contribution of the overall architecture design to performance, as shown in Table \ref{tab6}. Experimental results demonstrate that incorporating the FCPG module consistently improves performance across all tasks, achieving an F1 score improvement of 0.32\% and an IoU improvement of 0.47\%. This validates that the frequency  prompt mechanism compensates for information loss during the encoding process in existing methods, while also confirming its crucial role in suppressing pseudo-changes. \par
\textbf{Frequency Prompt Change Modeling Mechanism.} To further validate the effectiveness of the frequency prompt modeling mechanism, we conducted an in-depth ablation analysis of the internal components of the frequency change prompt generator (FCPG), as shown in Table \ref{tab6}. Experimental results indicate that removing the spatial modulator (SPM) component resulted in a 0.51\% decrease in Pre score. This validates that the component effectively enhances the model's ability to capture fine-grained structures by improving the sensitivity of boundary region responses through spatial adaptive weighting within local receptive fields. Regarding frequency settings, replacing adaptive adjustment of high and low frequency settings with fixed thresholds (low-frequency 0.1, high-frequency 0.3) resulted in a 0.57\% decrease in Pre score, while using a single frequency threshold (0.1) caused a 1.2\% drop Pre score. This clearly demonstrates that the adaptive high-low frequency coordination mechanism not only achieves coarse localization of change regions through low-frequency information but also refines boundaries using high-frequency information. They form a detection process progressing from coarse to fine, jointly ensuring the final detection accuracy. Dissolution experiments demonstrate that the frequency change prompt generator modeling mechanism ensures precise recognition of local detail features and effective suppression of pseudo-changes through the synergistic interaction of two core components: adaptive frequency of high and low frequency settings, spatial modulator.\par

\begin{figure}[t]
	\centering
	\includegraphics[scale=0.40,trim=190pt 50pt 190pt 30pt,clip]{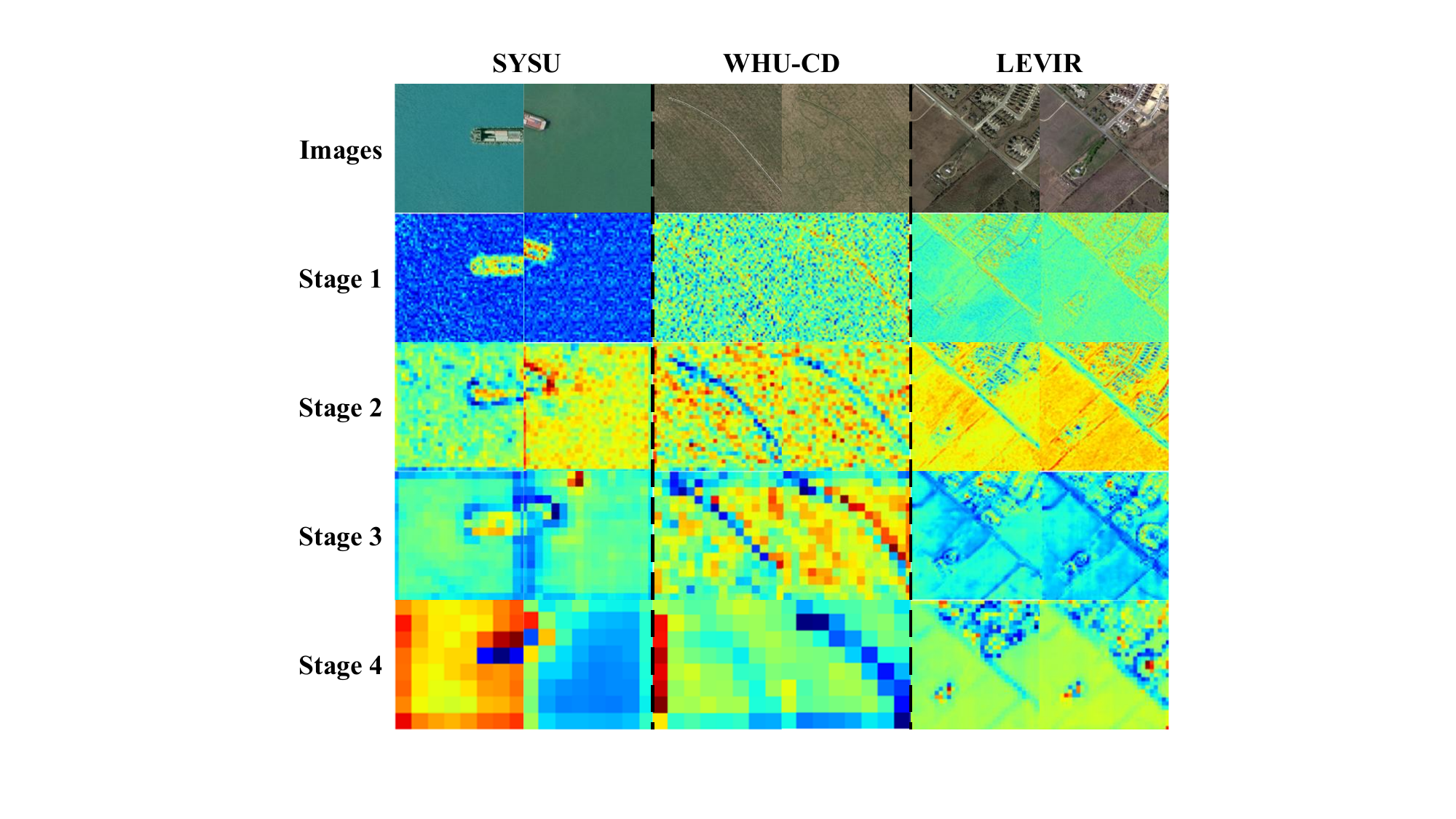}
	\caption{Visualization of features at different layers of UniRSCD across three datasets.}
	\label{fig3}
\end{figure}

\textbf{Bitemporal Concatenation Methods.} To validate the effectiveness of the bitemporal horizontal concatenation method, we conducted relevant experiments comparing the horizontal concatenation approach with the channel concatenation approach. We have removed the frequency change prompt generator from UniRSCD as an experimental model, named UniRS-noFCPG. Tables \ref{tab:13} and \ref{tab:14} compare the performance of UniRS-noFCPG framework in channel concatenation and horizontal concatenation on the SYSU and SECOND datasets. The results clearly demonstrate the superiority of horizontal concatenation, which exhibits both higher efficiency and greater stability. The channel connection approach primarily focuses on integrating channel information, while the distinction between changed and unchanged regions in bitemporal is characterized by their spatial variation. Unlike channel concatenation approach, horizontal concatenation approach connects two feature maps side-by-side, enabling UniRSCD to learn correlations between corresponding pixels across the bitemporal feature maps. Therefore, UniRSCD employs a horizontal connection method to link two temporal phases as input, enabling a more comprehensive capture of global spatial features.

\begin{table}[t]
	\centering
	\caption{Comparison of different concatenation methods on the SYSU dataset.}
	\label{tab:13}
	\tabcolsep 0.1cm 
	\begin{tabular}{lccccc c c c}
		\toprule
		\makecell[l]{Concatenation method} & Rec & Pre & F1 & IoU \\
		\midrule
		Channel concatenation & \textbf{80.95} & 86.33 & 83.13 & 70.68 \\
		Horizontal concatenation & 79.11 & \textbf{86.78} & \textbf{83.42} & \textbf{71.56} \\
		\bottomrule
	\end{tabular}
\end{table}

\begin{table}[t]
	\centering
	\caption{Comparison of different concatenation methods on the SECOND dataset. }
	\label{tab:14}
	\tabcolsep 0.1cm 
	\begin{tabular}{lccccc c c c}
		\toprule
		\makecell[l]{Concatenation method} & OA & F1 & mIoU & SeK37 \\
		\midrule
		Channel concatenation & 84.54 & 50.16 & 69.99 & 17.00 \\
		Horizontal concatenation & \textbf{86.38} & \textbf{56.14} & \textbf{72.60} & \textbf{22.48} \\
		\bottomrule
	\end{tabular}
\end{table}

\textbf{Visualization.} We randomly sampled data from the SYSU, LEVIR-CD, and WHU-CD datasets and performed visual analysis on the feature maps processed by the frequency change prompt module across the encoder's four stages as shown in Figure \ref{fig3}. The following key conclusions can be drawn from the feature maps: (1) As network depth increases, the response values in the changed regions of the feature maps processed by the frequency change prompt generator module gradually strengthen, while the responses in the unchanged regions and pseudo-changed regions are effectively suppressed. (2) The state space model and frequency change prompt module jointly serve to incorporate both global and local information, eliminating the need for a specific encoder to compensate for encoding process losses, thereby laying the foundation for a unified framework.
\section{Conclusion}

This paper proposes a novel unified change detection paradigm named UniRSCD. The framework integrates state space models with frequency change prompts into a unified encoder, enabling the model to simultaneously capture both global and local information. This effectively addresses the information loss inherent in traditional encoding processes. Furthermore, we introduce a structurally simple decoder alongside a multi-task adaptive prediction head, integrating tasks with varying output granularities into a single architecture. Comprehensive experiments validate UniRSCD's generality, showing that it achieves state-of-the-art results on a diverse set of change detection tasks. We hope that our work will serve diverse change detection tasks and inspire further research in related field.
{
    \small
    \bibliographystyle{ieeenat_fullname}
    \bibliography{main}
}


\end{document}